\documentclass[10pt,journal,letterpaper,compsoc]{IEEEtran}
\newcommand{\comment}[1]{}

\newcommand{\david}[1]{\textcolor{blue}{\textbf{[David: #1]}}}

\newcommand{\lookups}[0]{lookups}           
\newcommand{\Q}[0]{{\bf g}}           
\newcommand{\B}[0]{q}           
\newcommand{\s}[0]{s}           
\newcommand{\R}[0]{r}           
\newcommand{\N}[0]{n}           
\newcommand{\M}[0]{m}           

\usepackage{xspace}
\usepackage{amsfonts}
\usepackage{amsmath}

\newcommand{\bg}{{\mathbf g}}
\newcommand{\bh}{{\mathbf h}}

\newcommand{\Codeset}{\mathcal H}

\newcommand{\logtwo}{{{\log}_2}}

\newsavebox{\savepar}

\def\th#1{#1^\text{th}}

\def\ie{{\em i.e.,}}
\def\eg{{\em e.g.,}}
\def\vs{{\em vs.}}

\def\knn{$k$NN}
\def\nn#1{{#1-NN}}

\def\supj{{(j)}}

\newcommand{\secref}[1]{Section~\ref{#1}} 
\newcommand{\tableref}[1]{Table~\ref{#1}} 
\newcommand{\figref}[1]{Figure~\ref{#1}} 
\newcommand{\algref}[1]{Algorithm~\ref{#1}}

\ifCLASSOPTIONcompsoc
\else
\fi
\ifCLASSINFOpdf
  \usepackage[pdftex]{graphicx}
\else
\fi
\usepackage{algorithmic,algorithm}
\usepackage{multirow}
\usepackage{color}
\usepackage{pgfplots}
\usepackage{tikz}
\begin{document}
%
\title{Fast Exact Search in Hamming Space\\ with Multi-Index Hashing}
%
%
%
%

\author{Mohammad Norouzi, 
        Ali Punjani, 
        David J.\ Fleet, \\
        {\small \{norouzi, alipunjani, fleet\}@cs.toronto.edu}
}

%
%

\markboth{}%
{Shell \MakeLowercase{\textit{et al.}}: Bare Demo of IEEEtran.cls for Computer Society Journals}
%



\IEEEcompsoctitleabstractindextext{%
\begin{abstract}

There is growing interest in representing image data and feature descriptors using compact binary codes for fast near neighbor search. Although binary codes are motivated by their use as direct indices (addresses) into a hash table, codes longer than 32 bits are not being used as such, as it was thought to be ineffective. We introduce a rigorous way to build multiple hash tables on binary code substrings that enables exact k-nearest neighbor search in Hamming space. The approach is storage efficient and straight-forward to implement. Theoretical analysis shows that the algorithm exhibits sub-linear run-time behavior for uniformly distributed codes. Empirical results show dramatic speedups over a linear scan baseline for datasets of up to one billion codes of 64, 128, or 256 bits.

\end{abstract}


}

\maketitle

\IEEEdisplaynotcompsoctitleabstractindextext

%

\IEEEpeerreviewmaketitle

\section{Introduction}

\IEEEPARstart{T}here has been growing interest in representing image
data and feature descriptors in terms of compact binary codes, often
to facilitate fast near neighbor search and feature matching in vision
applications (\eg~\cite{AlahiCVPR12, CalonderECCV10,
  ShakhnarovichVD03,StrechaPAMI12, TorralbaCVPR08,KuettelGFECCV12}).
Binary codes are storage efficient and comparisons require just a
small number of machine instructions.  Millions of binary codes can be
compared to a query in less than a second.  But the most compelling
reason for binary codes, and discrete codes in general, is their use
as direct indices (addresses) into a hash table, yielding a dramatic
increase in search speed compared to an exhaustive linear scan
(\eg~\cite{WeissTF08,SalakhutdinovH09,NorouziICML11}).

Nevertheless, using binary codes as direct hash indices
is not necessarily efficient.  To find near neighbors one needs to
examine all hash table entries (or {\em buckets}) within some Hamming
ball around the query.  The problem is that the number of such buckets
grows near-exponentially with the search radius.
Even with a small search radius, the number of buckets to examine is 
often  larger than the number of items in the database, and hence slower 
than linear scan.  Recent papers on binary codes mention the use of hash 
tables, but resort to linear scan when codes are longer than $32$ bits 
(\eg~\cite{TorralbaCVPR08, SalakhutdinovH09, KulisD09,NorouziICML11}).
Not surprisingly, code lengths are often significantly 
longer than 32 bits in order to achieve satisfactory retrieval performance
(\eg~see Fig.\ \ref{fig:recall_1M_1B}).

This paper presents a new algorithm for exact $k$-nearest neighbor
(\knn) search on binary codes that is dramatically faster than
exhaustive linear scan. This has been an open problem since the
introduction of hashing techniques with binary codes.  Our new
multi-index hashing algorithm exhibits sub-linear search times,
while it is storage efficient, and straightforward to
implement.  Empirically, on databases of up to 1B codes we find that
multi-index hashing is hundreds of times faster than linear scan.
Extrapolation suggests that the speedup gain grows quickly with
database size beyond 1B codes.

\subsection{Background: Problem and Related Work}

Nearest neighbor (NN) search on binary codes is used for image search
\cite{RaginskyNIPS09,TorralbaCVPR08,WeissTF08}, matching local
features \cite{AlahiCVPR12,CalonderECCV10,JegouECCV08,StrechaPAMI12},
image classification \cite{BergamoNIPS11}, object segmentation
\cite{KuettelGFECCV12}, and parameter estimation
\cite{ShakhnarovichVD03}.  Sometimes the binary codes are generated
directly as feature descriptors for images or image patches, such as
BRIEF~\cite{CalonderECCV10}, FREAK~\cite{AlahiCVPR12},
\cite{BergamoNIPS11}, or \cite{TrzcinskiNIPS12}, and sometimes binary
corpora are generated by discrete similarity-preserving mappings from
high-dimensional data.  Most such mappings are designed to preserve
Euclidean distance (\eg~\cite{GongLazebnikCVPR11,KulisD09,
  RaginskyNIPS09, StrechaPAMI12, WeissTF08}).  Others focus on
semantic similarity (\eg~\cite{NorouziICML11, ShakhnarovichVD03,
  SalakhutdinovH09, TorralbaCVPR08, NorouziNIPS12, RastegariECCV2012,
  LiuCVPR2012}).  Our concern in this paper is not the algorithm used
to generate the codes, but rather with fast search in Hamming
space.\footnote{ There do exist several other promising approaches to
  fast approximate NN search on large real-valued image features
  (\eg~\cite{AlyBMVC11, jegouPAMI11, NorouziPF12, MujaICCVTA09,
    BabenkoCVPR12}).  Nevertheless, we restrict our attention in this
  paper to compact binary codes and exact search.}

We address two related search problems in Hamming space.  Given a
dataset of binary codes, $\Codeset$ $\equiv \{ \bh_i \}_{i=1}^n$, the
first problem is to find the $k$ codes in $\Codeset$ that are closest
in Hamming distance to a given query, \ie~\knn{} search in Hamming
distance.  The \nn{$1$} problem in Hamming space was called the {\em Best
  Match} problem by Minsky and Papert~\cite{MinskyPapert1969}.  They
observed that there are no obvious approaches significantly better
than exhaustive search, and asked whether such approaches might exist.

\begin{figure}

\begin{center}
\includegraphics[width=2.2in]{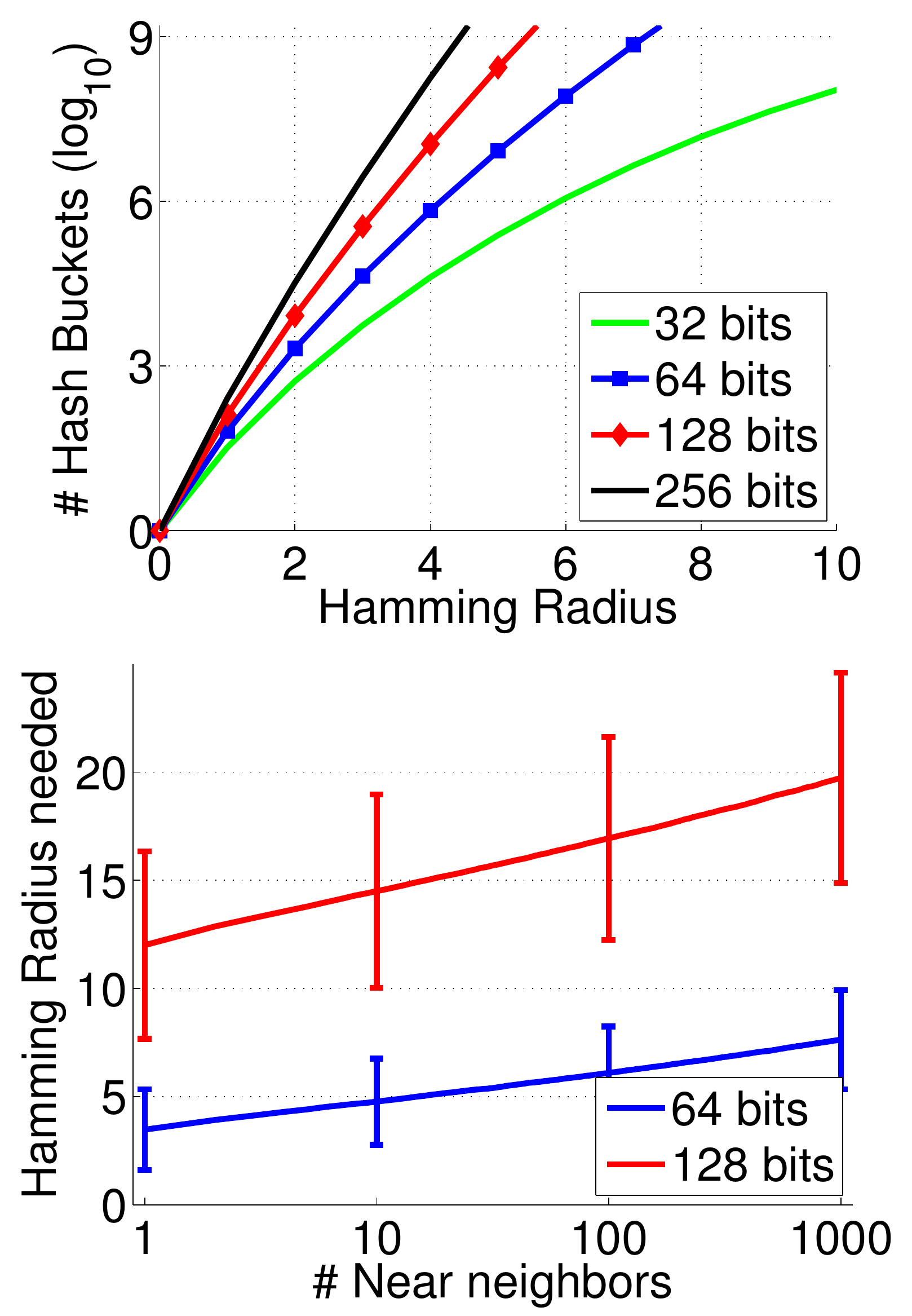}
\end{center}

\vspace*{-0.2cm}
\caption{
\label{fig:growth-n-choose-k}
(Top) Curves show the ($\log_{10}$) number of distinct hash table
indices (buckets) within a Hamming ball of radius $\R$, for different
code lengths.  With 64-bit codes there are about 1B buckets within a
Hamming ball with a 7-bit radius.  Hence with fewer than 1B database
items, and a search radius of 7 or more, a hash table would be less
efficient than linear scan.  Using hash tables with 128-bit codes is
prohibitive for radii larger than 6.  (Bottom) This plot shows the 
expected search radius required for \nn{$k$} search as a function of $k$, 
based on a dataset of 1B SIFT descriptors.  Binary codes with 64 
and 128 bits were obtained by random projections (LSH) from the SIFT
descriptors \cite{jegouTDA11}.  Standard deviation bars help show that
large search radii are often required.
}
\vspace*{-0.2cm}
\end{figure}


The second problem is to find all codes in a dataset $\Codeset$ that
are within a fixed Hamming distance of a query, sometimes called the
{\em Approximate Query} problem \cite{GreenePY94}, or {\em Point
  Location in Equal Balls} (PLEB)~\cite{IndykM98}.  A binary code is
an {\em $\R$-neighbor} of a query code, denoted $\Q$, if it differs
from $\Q$ in $\R$ bits or less. We define the {\em $\R$-neighbor
  search problem} as: find all $\R$-neighbors of a query $\Q$ from
$\Codeset$.

One way to tackle $\R$-neighbor search is to use a hash table
populated with the binary codes $\bh \in \Codeset$, and examine all
hash {\em buckets} whose indices are within $\R$ bits of a query
$\Q$ (\eg\, \cite{TorralbaCVPR08}).  For binary codes of $\B$ bits,
the number of distinct hash buckets to examine is
\begin{equation}
L(\B,\R) ~ =~  \sum_{z=0}^{\R} {\B \choose z} \, .
\end{equation}
As shown in Fig.~\ref{fig:growth-n-choose-k} (top), $L(\B,\R)$ grows 
very rapidly with $\R$.  Thus, this approach is only practical for small 
radii or short code lengths.  Some vision applications restrict search 
to exact matches (\ie~$r=0$) or a small search radius
(\eg~\cite{HeEtAlCVPR11,WangICML10-SPLH} ), but in most cases 
of interest the desired search radius is larger than is currently 
feasible (\eg~see Fig.\ \ref{fig:growth-n-choose-k} (bottom)).

Our work is inspired in part by the multi-index hashing results of
Greene, Parnas, and Yao~\cite{GreenePY94}.  Building on the classical
Turan problem for hypergraphs, they construct a set of over-lapping
binary substrings such that any two codes that differ by at most $\R$
bits are guaranteed to be identical in at least one of the constructed
substrings.  Accordingly, they propose an exact method for finding all
$\R$-neighbors of a query using multiple hash tables, one for each
substring.  At query time, candidate $\R$-neighbors are found by using
query substrings as indices into their corresponding hash tables.  As
explained below, while run-time efficient, the main drawback of their
approach is the prohibitive storage required for the requisite number
of hash tables.  By comparison, the method we propose requires much
less storage, and is only marginally slower in search performance.

While we focus on exact search, there also exist algorithms for
finding {\em approximate} $r$-neighbors ($\epsilon$-PLEB), or
approximate nearest neighbors ($\epsilon$-NN) in Hamming distance.
One example is Hamming Locality Sensitive Hashing~\cite{IndykM98,
GionisIM99}, which aims to solve the $(\R,\epsilon)$-neighbors 
decision problem: determine whether there exists a binary code 
$\bh \in \Codeset$ such that $\lVert \bh - \Q \rVert_H \le \R$, 
or whether all codes in $\Codeset$ differ from $\Q$ 
in $(1+\epsilon)\R$ bits or more. Approximate methods are 
interesting, and the approach below could be made faster by 
allowing misses.  Nonetheless, this paper will focus on
the {\em exact} search problem.

This paper proposes a data-structure that applies to both \knn{} and
$\R$-neighbor search in Hamming space.  We prove that for uniformly
distributed binary codes of $\B$ bits, and a search radius of $\R$
bits when $\R/\B$ is small, our query time is sub-linear in the size
of dataset.  We also demonstrate impressive performance on real-world
datasets.  To our knowledge this is the first practical data-structure
solving exact \knn{} in Hamming distance.

Section 2 describes a multi-index hashing algorithm for $\R$-neighbor 
search in Hamming space, followed by run-time and memory analysis 
in \secref{sec:mih-analysis}.  Section \secref{sec:k-nearest-neighbor} 
describes our algorithm for $k$-nearest neighbor search, and 
Section \secref{sec:expt} reports results on empirical datasets.

\section{Multi-Index Hashing}
\label{mih}

Our approach is called multi-index hashing, as binary codes from the
  database are indexed $\M$ times into $\M$ different hash tables,
  based on $\M$ disjoint substrings.
Given a query code, entries that fall
{\em close} to the query in at least one such substring are considered 
{\em neighbor candidates}.  Candidates are then checked for validity 
using the entire binary code, to remove any non-$\R$-neighbors. 
To be practical for large-scale datasets, the substrings must be 
chosen so that the set of candidates is small, and storage requirements 
are reasonable.  We also require that all true neighbors will be found.

The key idea here stems from the fact that, with $\N$ binary codes of
$\B$ bits, the vast majority of the $2^\B$ possible buckets in a full
hash table will be empty, since $2^\B \gg \N$.  It seems expensive to
examine all $L(\B,\R)$ buckets within $\R$ bits of a query, since most
of them contain no items.  Instead, we merge many buckets together
(most of which are empty) by marginalizing over different dimensions
of the Hamming space.  We do this by creating hash tables on 
{\em substrings} of the binary codes.  The distribution of the code 
substring comprising the first $s$ bits is the outcome of marginalizing 
the distribution of binary codes over the last $\B-s$ bits.  As such,
a given bucket of the substring hash table includes all codes with the 
same first $s$ bits, but having any of the $2^{(\B - s)}$ values for the 
remaining $\B-s$ bits.  Unfortunately these larger buckets are not 
restricted to the Hamming volume of interest around
the query.  Hence not all items in the merged buckets are
$\R$-neighbors of the query, so we then need to cull any candidate 
that is not a true $\R$-neighbor.



\subsection{Substring Search Radii}

In more detail, each binary code $\bh$, comprising $\B$ bits, is
partitioned into $\M$ disjoint substrings, $\bh^{(1)}, \ldots,
\bh^{(m)}$, each of length $\B / \M$
bits.  For convenience in what follows, we assume that $\B$
is divisible\footnote{When $\B$ is not divisible by $\M$, we
    use substrings of different lengths with either $\lfloor
    \frac{\B}{\M} \rfloor$ or $\lceil \frac{\B}{\M} \rceil$ bits,
    i.e., differing by at most 1 bit.
} by
$\M$, and that the substrings comprise contiguous bits.  The key idea
rests on the following statement: When two binary codes $\bh$ and
$\bg$ differ by $\R$ bits or less, then, in at least one of their $\M$
substrings they must differ by at most $\lfloor \R / \M \rfloor$
bits. This leads to the first proposition:

\vspace*{0.1cm}
\noindent
{\em Proposition 1:}
If $\lVert \bh - \bg \rVert_H \le \R$, where $\lVert \bh - \bg
\rVert_H$ denotes the Hamming distance between $\bh$ and $\bg$, then
\begin{eqnarray}
\exists~ 1 \le z \le m~~~s.t.~~~\lVert \, \bh^{(z)} - \bg^{(z)} \rVert_H ~\le~  \R' ~, 
\label{eqn1}
\end{eqnarray}
where $\R' = \left\lfloor \R / \M \right\rfloor$.\\
Proof of Proposition 1 follows straightforwardly from the Pigeonhole
Principle.  That is, suppose that the Hamming distance between each of the $\M$
substrings is strictly greater than $\R'$.  Then, 
$\lVert \bh -\bg\rVert_H \ge \M\, (\R'+1)$. Clearly, $ \M \,(\R'+1) > \R$, since 
$\R = \M\, \R' + a$ for some $a$ where $0\le a < \M$, which contradicts the 
premise.

The significance of Proposition 1 derives from the fact
that the substrings have only $\B / \M$ bits, and that the required
search radius in each substring is just $\R' = \lfloor \R / \M
\rfloor$.  For example, if $\bh$ and $\bg$ differ by $3$ bits or less,
and $\M = 4$, at least one of the 4 substrings must be identical.  If
they differ by at most $7$ bits, then in at least one substring they
differ by no more than $1$ bit; \ie~we can search a Hamming radius of
$7$ bits by searching a radius of $1$ bit on each of 4 substrings.
More generally, instead of examining $L(\B, \R)$ hash buckets, it
suffices to examine $L(\B/\M, r' )$ buckets in each of $\M$ substring
hash tables, \ie~ a total of $\M\,L(\B/\M, r' )$ buckets.


While it suffices to examine all buckets within a radius of $\R'$ in
all $\M$ hash tables, we next show that it is not always necessary.
Rather, it is often possible to use a radius of just $\R'-1$ in some
of the $\M$ substring hash tables while still guaranteeing that all
$\R$-neighbors of $\Q$ will be found.  In particular, with $\R = \M\R'
+ a$, where $0\le a<\M$, to find any item within a radius of $\R$ on
$\B$-bit codes, it suffices to search $a+1$ substring hash tables to a
radius of $\R'$, and the remaining $m-(a+1)$ substring hash tables up
to a radius of $\R'-1$.  Without loss of generality, since there is no
order to the substring hash tables, we search the first $a+1$ hash
tables with radius $\R'$, and all remaining hash tables with radius
$\R'-1$.

\vspace*{0.1cm}
\noindent
{\em Proposition 2:}
If $ || {\bf h} -{\Q} ||_H \le \R = \M \R' + a$, then
\begin{subequations}
\label{eqn:disj}
\begin{align}
\exists~  1 \le z \le a+1 ~~~ s.t. ~~~&
\lVert\, {\bf h}^{(z)} - \Q^{(z)} \rVert_H \le \R'  \label{eqn:disj1} \\
\mbox{OR} \nonumber \\
\exists~ a+1 < z \le \M ~~~ s.t. ~~~&
\lVert\,{\bf h}^{(z)} - \Q^{(z)}  \rVert_H \le \R'-1  \label{eqn:disj2}\,.
\end{align}
\end{subequations}
To prove Proposition 2, we show that when \eqref{eqn:disj1} is false,
(\ref{eqn:disj2}) must be true.  If \eqref{eqn:disj1} is false, then
it must be that $a<\M-1$, since otherwise $a = \M-1$, in which case
\eqref{eqn:disj1} and Proposition 1 are equivalent.  If
(\ref{eqn:disj1}) is false, it also follows that ${\bf h}$ and $\Q$
differ in each of their first $a+1$ substrings by $\R'+1$ or more
bits.  Thus, the total number of bits that differ in the first $a+1$
substrings is at least $(a\!+\!1)(\R'\!+\!1)$.  Because $|| {\bf h}
-{\Q} ||_H\le \R$, it also follows that the total number of bits that
differ in the remaining $\M-(a\!+\!1)$ substrings is at most $\R -
(a\!+\!1)(\R'\!+\!1)$.  Then, using Proposition 1, the maximum search
radius required in each of the remaining $m-(a+1)$ substring hash
tables is
\begin{eqnarray}
\left\lfloor \frac{ \R - (a\!+\!1)(\R'\!+\!1) }{\M - (a\!+\!1)} \right\rfloor 
&\!\! = \! & 
\left\lfloor \frac{ \M\R' + a - (a\!+\!1)\R'  - (a\!+\!1) }{\M -(a\!+\!1)} \right\rfloor 
\nonumber \\
&\!\! = \! & 
\left \lfloor \R' - \frac{ 1}{\M - (a\!+\!1)} \right \rfloor 
\nonumber \\
&\!\! = \! & \R' - 1 ~,
\end{eqnarray}
and hence Proposition 2 is true.
Because of the near exponential growth in the number of buckets 
for large search radii, the smaller substring search radius 
required by Proposition 2 is significant.

A special case of Proposition 2 is when $\R < \M$, 
hence $\R' = 0$ and $a = \R$. In this case, it suffices to 
search $\R + 1$ substring hash tables for a radius of $\R' = 0$ 
(i.e., exact matches), and the remaining $\M - (\R+1)$  substring 
hash tables can be ignored.
Clearly, if a code does not match exactly with a query in any 
of the selected $\R+1$ substrings, then the code must differ 
from the query in at least $\R+1$ bits.

\subsection{Multi-Index Hashing for $\R$-neighbor Search}
\label{sec:mih-knn}

In a pre-processing step, given a dataset of binary codes, one hash
table is built for each of the $\M$ substrings, as outlined in
\algref{alg:pop-mih}. At query time, given a query $\Q$ with
substrings $ \{ {\Q}^{(j)} \}_{j=1}^{\M}$, we search the $j^\text{th}$
substring hash table for entries that are within a Hamming distance of
$\lfloor r / m \rfloor$ or $\lfloor r / m \rfloor \, -\,1$ of
$\Q^{(j)}$, as prescribed by~(\ref{eqn:disj}).  By doing so we obtain
a set of candidates from the $j^\text{th}$ substring hash table,
denoted $\mathcal{N}_j(\Q)$.  According to the propositions above, the
union of the $\M$ sets, $\mathcal{N}(\Q) = \bigcup_j
\mathcal{N}_j(\Q)$, is necessarily a superset of the $r$-neighbors of
$\Q$.  The last step of the algorithm computes the full Hamming
distance between $\Q$ and each candidate in $\mathcal{N}(\Q)$,
retaining only those codes that are true $r$-neighbors of
$\Q$. \algref{alg:mih-rneighbor} outlines the $\R$-neighbor retrieval
procedure for a query $\Q$.

\begin{algorithm}[t]
\caption{Building $\M$ substring hash tables.}
\label{alg:pop-mih}
\begin{algorithmic}
\STATE Binary code dataset: $ \Codeset = \{ \bh_i \}_{i=1}^{n}$ 
\FOR{$j=1$ to $m$}
\STATE Initialize $j^\text{th}$ hash table
\FOR{$i=1$ to $n$}
\STATE Insert $\bh_i^\supj$ into $j^\text{th}$ hash table
\ENDFOR
\ENDFOR
\end{algorithmic}
\end{algorithm}

The search cost depends on the number of {\em lookups} (\ie~the
number of buckets examined), and the number of candidates tested.  
Not surprisingly there is a natural trade-off between them.  
With a large number of lookups one can minimize the number of
extraneous candidates.  By merging many buckets to reduce the number
of lookups, one obtains a large number of candidates to test.  In the
extreme case with $\M = \B$, substrings are 1 bit long, so we can
expect the candidate set to include almost the entire database.

Note that the idea of building multiple hash tables is not novel in
itself (\eg\ see \cite{GreenePY94, IndykM98}). However previous work
relied heavily on exact matches in substrings.  Relaxing this
constraint is what leads to a more effective algorithm, especially 
in terms of the storage requirement.

\section{Performance Analysis}
\label{sec:mih-analysis}

We next develop an analytical model of search performance to help 
address two key questions: (1) How does search cost depend on substring 
length, and hence the number of substrings? (2) How do run-time and storage
complexity depend on database size, code length, and search radius?

To help answer these questions we exploit a well-known bound on the
sum of binomial coefficients \cite{FlumGrohe2006}; \ie~for any 
$0< \epsilon \le \frac{1}{2}$ and $\eta \ge 1$.
\begin{equation}
\sum_{\kappa=0}^{\lfloor \epsilon \, \eta \rfloor} {\eta \choose \kappa} 
~\le~ 2 ^ {H(\epsilon)\, \eta }~,
\label{eq:bound-sum-binomial}
\end{equation}
where $H(\epsilon) \equiv -\epsilon \log_2 \epsilon -(1\! -\!\epsilon)
\log_2(1\!-\!\epsilon)$ is the entropy of a Bernoulli distribution
with probability $\epsilon$.

In what follows, $\N$ continues to denote the number of $\B$-bit
database codes, and $\R$ is the Hamming search radius.  Let $\M$
denote the number of hash tables, and let $\s$ denote the substring
length $\s = \B / \M$.  Hence, the maximum substring search radius
becomes $\R' = \lfloor\R / \M\rfloor =\lfloor\s\,\R / \B\rfloor$.  As
above, for the sake of model simplicity, we assume $\B$ is divisible
by $\M$.

We begin by formulating an upper bound on the number of lookups.
First, the number of lookups in \algref{alg:mih-rneighbor} is bounded
above by the product of $\M$, the number of substring hash tables, and
the number of hash buckets within a radius of $\lfloor\s\,\R /
\B\rfloor$ on substrings of length $\s$ bits.  Accordingly, using
(\ref{eq:bound-sum-binomial}), if the search radius is less than half
the code length, $\R \le \B/2 \,$, then the total number of lookups is
given by
\begin{equation}
\lookups(s) ~=~ 
\frac{~\B~}{\s}\!\sum_{z=0}^{\lfloor \s\,\R / \B \rfloor} \! {\s \choose z} 
~\le~
\frac{~\B~}{\s}~2^{H(\R/\B)\s}~.
\label{eq:bound-lookups}
\end{equation}
Clearly, as we decrease the substring length $\s$, thereby increasing 
the number of substrings $\M$, exponentially fewer lookups are needed. 

To analyze the expected number of candidates per bucket, we consider
the case in which the $\N$ binary codes are uniformly distributed over
the Hamming space.  In this case, for a substring of $\s$ bits, for
which a substring hash table has $2^\s$ buckets, the expected number
of items per bucket is $\N / 2^s$.  The expected size of the candidate
set therefore equals the number of lookups times $\N / 2^s$.

The total search cost per query is the cost for lookups plus the cost
for candidate tests.  While these costs will vary with the code length
$\B$ and the way the hash tables are implemented, empirically 
we find that, to a reasonable approximation, the costs of a lookup 
and a candidate test are similar (when $\B \le 256$).  Accordingly, 
we model the total search cost per query, for retrieving all 
$\R$-neighbors, in units of the time required for a single lookup, as
\begin{eqnarray}
cost(\s) &=& 
\left( 1 + \frac{\N}{2^\s} \right)\frac{~\B~}{\s}
\!\sum_{z=0}^{\lfloor \s\R / \B \rfloor} \! {\s \choose z}~,
\label{eq:cost-for-retrieval}  \\
&\le&
\left( 1 + \frac{\N}{2^\s} \right)
\frac{~\B~}{\s}~2^{H(\R/\B)\s}~.
\label{eq:costbound} 
\end{eqnarray}

In practice, database codes will not be uniformly
distributed, nor are uniformly distributed codes ideal for multi-index
hashing.  Indeed, the cost of search with uniformly distributed codes
is relatively high since the search radius increases as the density of
codes decreases.  Rather, the uniform distribution is primarily a
mathematical convenience that facilitates the analysis of run-time,
thereby providing some insight into the effectiveness of the approach
and how one might choose an effective substring length.

\begin{algorithm}[t]
\caption{$\R$-Neighbor Search for Query $\Q\, $.}
\label{alg:mih-rneighbor}
\begin{algorithmic}
\STATE Query substrings: $ \{ {\Q}^\supj \}_{j=1}^{\M}$ 
\STATE Substring radius: $r' = \lfloor \R / \M \rfloor$, and $a = \R - \M\R'$
\FOR{$j=1$ to $a+1$}
\STATE Lookup $\R'$-neighbors of $\Q^\supj$ from $j^{th}$ hash table
\ENDFOR
\FOR{$j=a+2$ to $\M$}
\STATE Lookup ($\R' $-$1$)-neighbors of $\Q^\supj$ from $j^{th}$ hash table
\ENDFOR
\STATE Remove all non $\R$-neighbors from the candidate set.
\end{algorithmic}
\end{algorithm}

\subsection{Choosing an Effective Substring Length}

As noted above in Sec.~\ref{sec:mih-knn}, finding a good 
substring length is central to the efficiency of multi-index hashing.  
When the substring length is too large or too small the approach 
will not be effective. In practice, an effective substring length 
for a given dataset can be determined by cross-validation. 
Nevertheless this can be expensive.

In the case of uniformly distributed codes, one can instead 
use the analytic cost model in~\eqref{eq:cost-for-retrieval} to find 
a near optimal substring length.  As discussed below, we find that
a substring length of $\s = \log_2 \N$ yields a near-optimal search 
cost.  Further, with non-uniformly distributed codes in benchmark
datasets, we find empirically that $\s = \log_2 \N$ is also a 
reasonable heuristic for choosing the substring length 
(\eg~see \tableref{tab:selected-m} below).
  
In more detail, to find a good substring length using the cost model 
above, assuming uniformly distributed binary codes, we first note that, 
dividing $cost(\s)$ in \eqref{eq:cost-for-retrieval} by $\B$ has no 
effect on the optimal $s$.  Accordingly, one can view the optimal $s$ 
as a function of two quantities, namely the number of items, $\N$, 
and the search ratio $\R/\B$.

\begin{figure}[t]
\begin{center}
\includegraphics[width=2.5in]{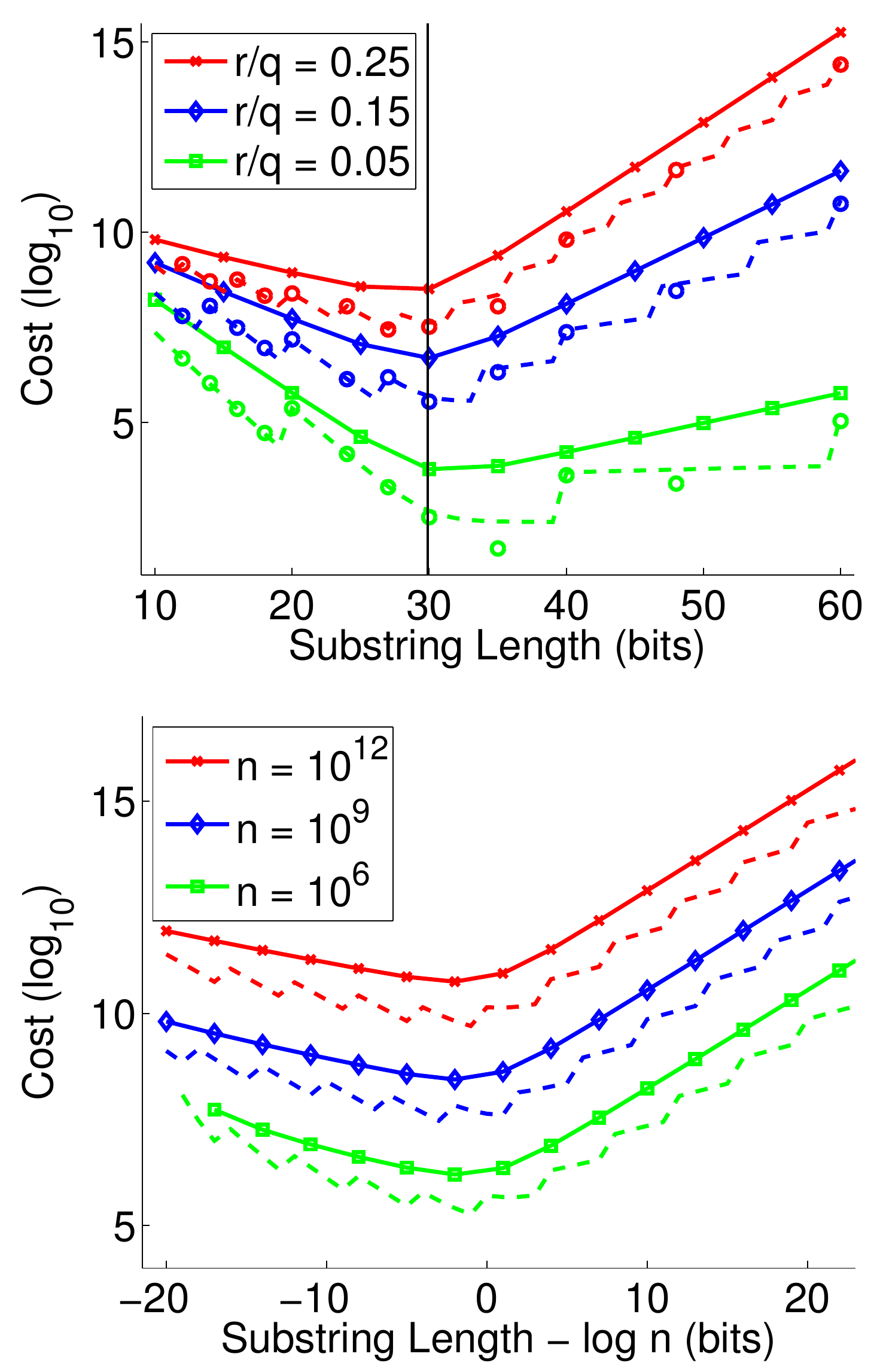} \\ 
\end{center}

\vspace*{-0.2cm}
\caption{
\label{costfig1}
Cost (\ref{eq:cost-for-retrieval}) and its upper bound (\ref{eq:costbound}) 
are shown as functions of substring length (using dashed and solid curves 
respectively).
The code length in all cases is $\B=240$ bits.  (Top) Cost for different
search radii, all for a database with $\N = 10^9$ codes.  Circles depict 
a more accurate cost measure, only for substring lengths that are integer 
divisors of $\B$, and with the more efficient indexing in \algref{alg:mih-knn}.
(Bottom) Three database sizes, all with a search radius of $\R = 0.25\,\B$.  
The minima are aligned when each curve is displaced horizontally 
by $-\log_2 \N$.
}
\end{figure}

Figure \ref{costfig1} plots  cost as a function of substring length $\s$,
for $240$-bit codes, different database sizes $\N$, and different search 
radii (expressed as a fraction of the code length $\B$). Dashed curves 
depict  $\,cost(\s) $  in (\ref{eq:cost-for-retrieval}) while solid curves 
of the same color depict the upper bound in  (\ref{eq:costbound}).
The tightness of the bound is evident in the plots, as are 
the quantization effects of the upper range of the sum in 
(\ref{eq:cost-for-retrieval}).
The small circles in Fig.\ \ref{costfig1} (top) depict cost when all
quantization effects are included, and hence it is only shown at 
substring lengths that are integer divisors of the code length.

Fig.\ \ref{costfig1} (top) shows cost for search radii equal to
$5\%$, $15\%$ and $25\%$ of the code length, with $\N\! =\! 10^9$ in
all cases.  One striking property of these curves is that the cost is
persistently minimal in the vicinity of $\s = \log_2 \N$, indicated by
the vertical line close to 30 bits.  This behavior is consistent
over a wide range of database sizes.  

Fig.\ \ref{costfig1} (bottom) shows the dependence of cost on $\s$ for
databases with $\N = 10^6$, $10^9$, and $10^{12}$, all with $\R/\B =
0.25$ and $\B = 128$ bits.  In this case we have laterally displaced
each curve by $-\log_2 \N$; notice how this aligns the minima close to
$0$.  These curves suggest that, over a wide range of conditions, cost
is minimal for $\s = \log_2 \N$.  For this choice of the substring length, 
the expected number of items per substring bucket, \ie~$ \N / 2^\s$, 
reduces to 1.  As a consequence, the number of lookups is equal to the 
expected number of candidates.  Interestingly, this choice of 
substring length is similar to that of Greene {\em et al.}~\cite{GreenePY94}.
A somewhat involved theoretical analysis based on Stirling's
approximation, omitted here, also suggests that as $\N$ goes to 
infinity, the optimal substring length converges asymptotically 
to $log_2 \N$.


\comment{{{
Under this assumption we aim to find $\s$ that minimizes the query cost. 
This gives some insight about the range of interesting values of $\s$ 
and their running time behavior. 
In Fig.~\ref{costfig1}, we plot $cost(\s)$ for a fixed $\N$ and
different values of $\R/\B$, and fixed $\R/\B$ and varying $\N$.  We
find out that $\s \approx \log_2{\N}$, which seems very intuitive,
because in this case the number of candidates and lookups become the
same, as each bucket in the substring hash tables will have an average
density of 1. 
}}}

\subsection{Run-Time Complexity}

Choosing $\s$ in the vicinity of $\log_2{\N}$ also permits
a simple characterization of retrieval run-time complexity, for
uniformly distributed binary codes.   When $s = \log_2{\N}$, the
upper bound on the number of lookups~(\ref{eq:bound-lookups}) also
becomes a bound on the number candidates.  In particular, if we
substitute $\log_2{\N}$ for $\s$ in (\ref{eq:costbound}), then we find
the following upper bound on the cost, now as a function of database
size, code length, and the search radius:
\begin{equation}
cost(\log_2 \N)  ~\le~ 2\, \frac{~\B~}{\log_2 \N}~\N^{H(\R/\B)}~.
\label{eq:bound-cost-logn}
\end{equation}

Thus, for a uniform distribution over binary codes, if we choose $\M$ 
such that $\s \approx \log_2 \N$, the expected query time complexity 
is $O(\B\, \N^{H(\R/\B)} / {\log_2\N})$.   For a small ratio of $\R / \B$ 
this is sub-linear in $\N$.  For example, if $\R/\B \le .11$, then 
$H(.11) < .5$, and the run-time complexity becomes 
$O(\B\, \sqrt{\N} / {\log_2\N})$.  That is, the search time 
increases with the square root of the database size when the search
radius is approximately 10\% of the code length.
For $\R/\B \le .06$, this becomes $O(\B\, \sqrt[3]{\N} / {\log_2\N})$.  
The time complexity with respect to $\B$ is not as important as 
that with respect to $\N$ since $\B$ is not expected to vary 
significantly in most applications.

\subsection{Storage Complexity}
\label{sec:storage-order}

The storage complexity of our multi-index hashing algorithm is
asymptotically optimal when $\lfloor \B / \log_2 \N \rfloor \le \M \le
\lceil\B / \log_2 \N \rceil$, as is suggested above.  To store the
full database of binary codes requires $O( \N \B )$ bits.  For each of
$\M$ hash tables, we also need to store $\N$ unique identifiers to the
database items.  This allows one to identify the retrieved items and
fetch their full codes; this requires an additional $O(\M \N \log_2 \N
)$ bits.  In sum, the storage required is $O(\N\B+\M \N \log_2 \N
)$. When $\M \approx \B / \log_2 \N$, this storage cost reduces to $O(
\N \B + \N \log_2 \N)$.  Here, the $\N \log_2 \N$ term does not cancel
as $\M \ge 1$, but in most interesting cases $\B > \log_2 \N$,
  so the $\N \log_2 \N$ term does not matter.

While the storage cost for our multi-index hashing algorithm is linear 
in $\N\B$, the related multi-index hashing algorithm of 
Greene {\em et al.}~\cite{GreenePY94} 
entails storage complexity that is super-linear in $\N$.  
To find all $\R$-neighbors, for a given search radius $\R$, 
they construct $\M = O(\R 2^{\s\R/\B})$ substrings of
length $\s$ bits per binary code. Their suggested substring length is
also $\s = \log_2 {\N}$, so the number of substring hash tables
becomes $\M = O(\R\N^{\R/\B})$, each of which requires $O(\N \log_2{\N})$ 
in storage.  As a consequence for
large values of $\N$, even with small $\R$, this technique requires a
prohibitive amount of memory to store the hash tables.


Our approach is more memory-efficient than that of~\cite{GreenePY94}
because we do not enforce exact equality in substring matching. In
essence, instead of creating all of the hash tables off-line, and then
having to store them, we flip bits of each substring at run-time and
implicitly create some of the substring hash tables on-line. This
increases run-time slightly, but greatly reduces storage
costs.


\comment{{{
Give or take miscellaneous extra storage required, suppose we assume $nd/8$ 
bytes to store full binary codes, and then an additional $mn\log{n} = nd/8$ 
for multiple hash tables for subsequences.  Then the cost is $2nd/8$.
If we also assume 4 bytes per feature to store geometric information (for 
NDID verification), we need another $4n$ bytes.  So the cost is $n(d/4 +4)$.
For code lengths $\B$ of $64$ and $128$ bits, and $n=10^9$ this gives 20 and 36 GBs.
For code lengths $\B$ of $64$ and $128$ bits, and $n=10^{12}$ this gives 20 and 36 TBs.
}}}

\section{$\mathbf k$-Nearest Neighbor Search}
\label{sec:k-nearest-neighbor}

To use the above multi-index hashing in practice, one must specify a
Hamming search radius $\R$.  For many tasks, the value of $\R$ is
chosen such that queries will, on average, retrieve $k$ near
neighbors.  Nevertheless, as expected, we find that for many hashing
techniques and different sources of visual data, the distribution of
binary codes is such that a single search radius for all queries will
not produce similar numbers of neighbors.

\begin{figure*}[t]
\vspace*{-.1cm}
\begin{center}
\includegraphics{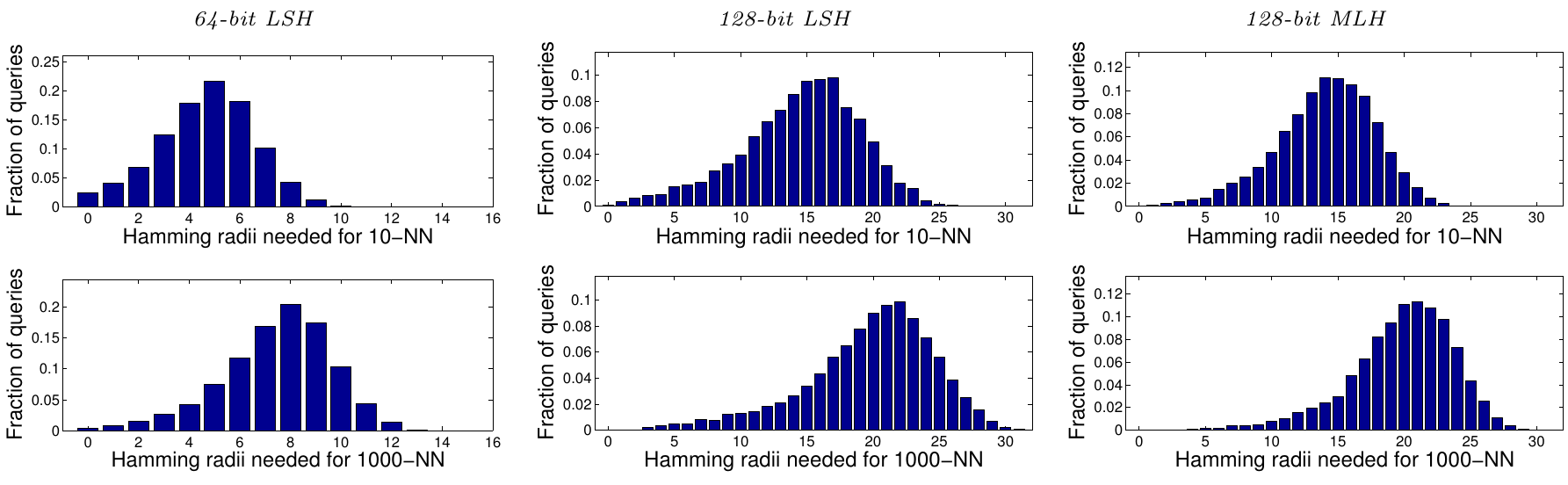}
\end{center}
\vspace*{-.1cm}

\caption{
\label{fig:radius-for-10-1000NN}
Shown are histograms of the search radii that are required to find $10$-NN 
and $1000$-NN, for $64$- and $128$-bit code from LSH \cite{AndoniIndyk08}, and 
$128$-bit codes from MLH \cite{NorouziICML11}, based on 1B SIFT
descriptors \cite{jegouTDA11}.
Clearly shown are the relatively large search radii required for both
the $10$-NN and the $1000$-NN tasks, as well as the increase in the radii
required when using $128$ bits versus $64$ bits.
}
\end{figure*}

\figref{fig:radius-for-10-1000NN} depicts empirical distributions of
search radii needed for $10$-NN and $1000$-NN on three sets of binary
codes obtained from 1B SIFT descriptors \cite{jegouTDA11,lowe04}.  In
all cases, for $64$- and $128$-bit codes, and for hash functions based
on LSH \cite{AndoniIndyk08} and MLH \cite{NorouziICML11}, there is a
substantial variance in the search radius.  This suggests that binary
codes are not uniformly distributed over the Hamming space.  As an
example, for $1000$-NN in $64$-bit LSH codes, about $10\%$ of the
queries require a search radius of $10$ bits or larger, while about
$10\%$ of the queries need a search radius of $5$ bits or
smaller.  Also evident from Fig.\ \ref{fig:radius-for-10-1000NN} is
the growth in the required search radius as one moves from $64$-bit
codes to $128$ bits, and from $10$-NN to $1000$-NN.

A fixed radius for all queries would produce too many neighbors for
some queries, and too few for others.  It is therefore more natural
for many tasks to fix the number of required neighbors, \ie~$k$, and
let the search radius depend on the query.  Fortunately, our
multi-index hashing algorithm is easily adapted to accommodate
query-dependent search radii.  

Given a query, one can progressively increase the Hamming search 
radius per substring, until a specified number of neighbors is found. 
For example, if one examines all $\R'$-neighbors of a query's substrings, 
from which more than $k$ candidates are found to be within a Hamming 
distance of $(\R'+1)\,\M-1$ bits (using the full codes for validation), 
then it is guaranteed that $k$-nearest neighbors have been found. 
Indeed, if all $k$NNs of a query $\Q$ differ from $\Q$ in $\R$ bits 
or less, then Propositions 1 and 2 above provide guanantees all such 
neighbors will be found if one searches the substring hash tables 
with the prescribed radii.

In our experiments, we follow this progressive increment of the search
radius until we can find \knn{} in the guaranteed neighborhood of a
query. This approach, outlined in \algref{alg:mih-knn}, is helpful because it
uses a query-specific search radius depending on the distribution of
codes in the neighborhood of the query.

\begin{algorithm}[t]
\caption{\knn{} Search with Query $\Q\,$.}
\label{alg:mih-knn}
\begin{algorithmic}
\STATE Query substrings: $ \{ {\Q}^{(i)} \}_{i=1}^{\M}$ 
\STATE Initialize sets: $N_d = \emptyset$, for $0 \le d \le \B$
\STATE Initialize integers: $\R' = 0, a = 0, \R = 0$
\REPEAT
\STATE Assertion: Full search radius should be $\R = \M\R' + a\, $.
\STATE Lookup buckets in the $(a\!+\!1)^{th}$ substring hash 
table that differ from $\Q^{(a\!+\! 1)}$ in exactly $\R'$ bits. 
\STATE For each candidate found, measure full Hamming distance, 
and add items with distance $d$ to $N_d$.
\STATE $a \leftarrow a+1$
\IF{ $a \ge \M$}  
\STATE $a \leftarrow 0$
\STATE $\R' \leftarrow \R' +1$ 
\ENDIF
\STATE $\R \leftarrow \R+1$ 
\UNTIL{ $\sum_{d=0}^{r-1} | N_d | \ge k\, $ (i.e., $k$ $r$-neighbors are found)}
\end{algorithmic}
\end{algorithm}

\section{Experiments}
\label{sec:expt}

Our implementation of multi-index hashing is available on-line
at \cite{mih}.  Experiments are run on two different architectures.
The first is a mid- to low-end $2.3$Ghz dual quad-core AMD Opteron
processor, with $2$MB of L2 cache, and $128$GB of RAM.  The second is
a high-end machine with a $2.9$Ghz dual quad-core Intel Xeon
processor, $20$MB of L2 cache, and $128G$B of RAM.  The difference in
the size of the L2 cache has a major impact on the run-time of linear
scan, since the effectiveness of linear scan depends greatly on L2
cache lines.  With roughly ten times the L2 cache, linear scan on the
Intel platform is roughly twice as fast as on the AMD machines.  By
comparison, multi-index hashing does not have a serial memory access
pattern and so the cache size does not have such a pronounced effect.
Actual run-times for multi-index hashing on the Intel and AMD
platforms are within 20\% of one another.

Both linear scan and multi-index hashing were implemented in C++ and
compiled with identical compiler flags.  To accommodate the large size
of memory footprint required for 1B codes, we used the libhugetlbfs
package and Linux Kernel $3.2.0$ to allow the use of $2$MB page sizes.
Further details about the implementations are given in
\secref{sec:details}. Despite the existence of multiple
cores, all experiments are run on a single core to simplify run-time
measurements.

\begin{figure}[t]
\begin{center}
\includegraphics[width=2.5in]{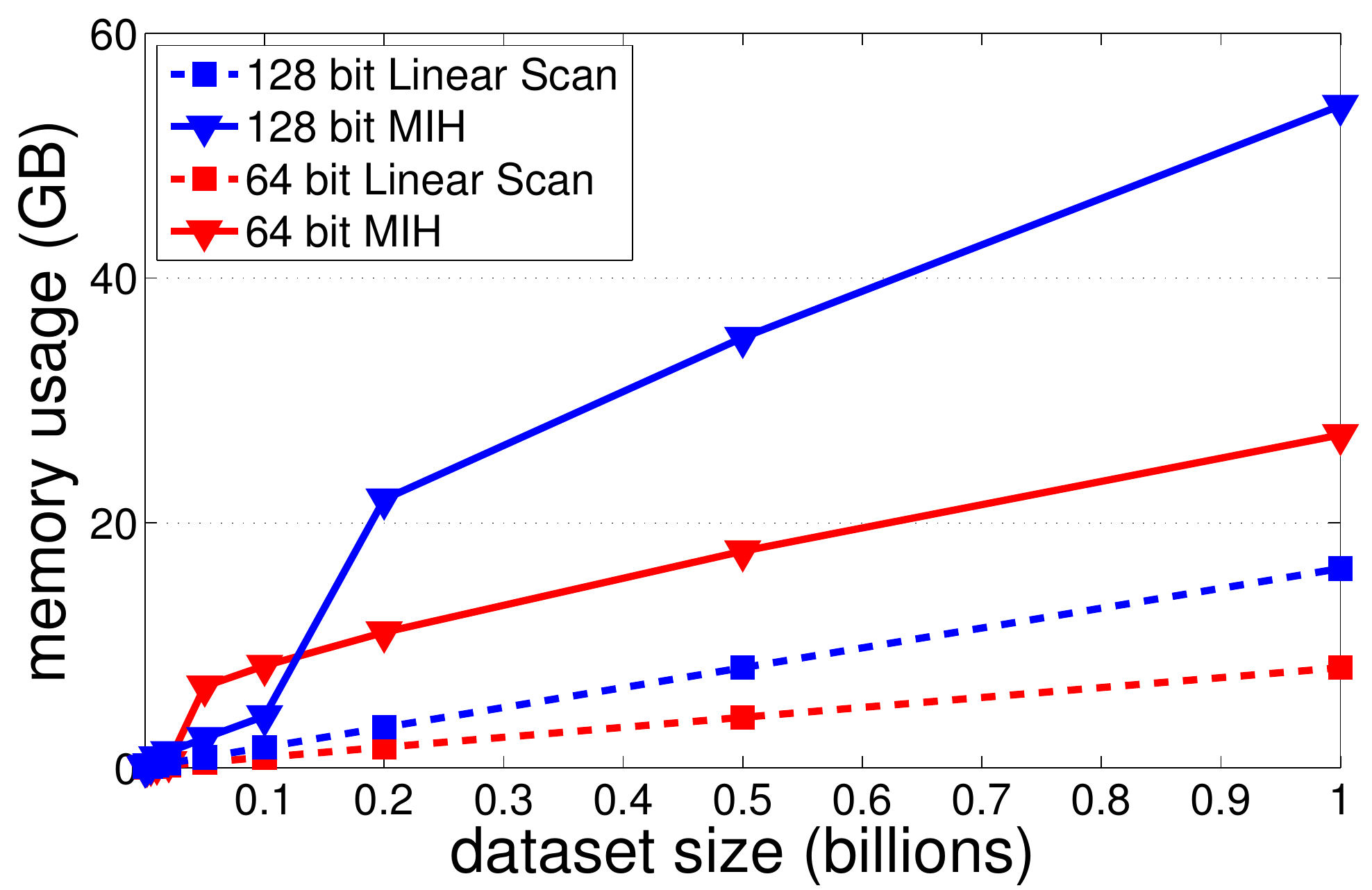} \\\
\end{center}
\caption{Memory footprint of our implementation of Multi-Index Hashing 
as a function of database size.  Note that the memory usage does 
not grow super-linearly with dataset size.  The memory usage is independent 
of the number of nearest neighbors requested.
}
\label{fig:storagePlot}
\end{figure}

The memory requirements for multi-index hashing are described in
detail in \secref{sec:details}.  We currently require approximately
$27\,$GB for multi-index hashing with 1B $64$-bit codes, and
approximately twice that for $128$-bit codes.
\figref{fig:storagePlot} shows how the memory footprint depends on
the database size for linear scan and multi-index hashing.  As
explained in the Sec.~\ref{sec:storage-order}, and demonstrated in
\figref{fig:storagePlot} the memory requirements of multi-index
hashing grow linearly in the database size, as does linear scan.
While we use a single computer in our experiments, one could implement
a distributed version of multi-index hashing on computers with much
less memory by placing each substring hash table on a separate
computer.

\subsection{Datasets}

\begin{figure}
\vspace*{-.1cm}
\begin{center}
\includegraphics[height=1.6in]{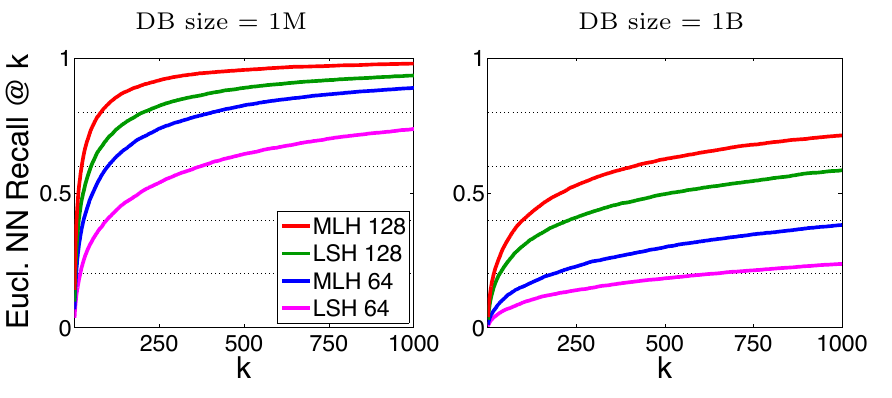}
\end{center}
\vspace*{-.4cm}
\caption{Recall rates for BIGANN dataset~\cite{jegouTDA11} ($1$M and $1$B
  subsets) obtained by \knn{} on $64$- and $128$-bit MLH and LSH codes.
}
\label{fig:recall_1M_1B}
\vspace*{-.1cm}
\end{figure}

We consider two well-known large-scale vision corpora: $80$M Gist 
descriptors from $80$ million tiny images~\cite{TorralbaFF08} and 
$1$B SIFT features from the BIGANN dataset~\cite{jegouTDA11}.  
SIFT vectors~\cite{lowe04} are $128$D
descriptors of local image structure in the vicinity of feature
points.  Gist features~\cite{OlivaIJCV01} extracted from $32
\times 32$ images capture global image structure in $384$D vectors.
These two feature types cover a spectrum of NN search problems in
vision from feature to image indexing.

We use two similarity-preserving mappings to create datasets of binary
codes, namely, binary angular Locality Sensitive Hashing
(LSH)~\cite{Charikar02}, and Minimal Loss Hashing
(MLH)~\cite{NorouziICML11,NorouziNIPS12}. LSH is considered a baseline
random projection method, closely related to cosine similarity.  MLH
is a state-of-the-art learning algorithm that, given a set of
similarity labels, optimizes a mapping by minimizing
a loss function over pairs or triplets of binary codes.

Both the $80$M Gist and $1$B SIFT corpora comprise three disjoint
sets, namely, a training set, a base set for populating the database,
and a test query set.  Using a random permutation, Gist descriptors
are divided into a training set with $300K$ items, a base set of $79$
million items, and a query set of size $10^4$.  The SIFT corpus comes
with $100$M for training, $10^9$ in the base set, and $10^4$ test
queries.

For LSH we subtract the mean, and pick a set of coefficients from the
standard normal density for a linear projection, followed by
quantization.  For MLH the training set is used to optimize 
hash function parameters~\cite{NorouziNIPS12}.  After learning is
complete, we remove the training data and use the resulting hash
function with the base set to create the database of binary codes.
With two image corpora (SIFT and Gist), up to three code lengths
($64$, $128$, and $256$ bits), and two hashing methods (LSH and MLH),
we obtain several datasets of binary codes with which to evaluate our
multi-index hashing algorithm. Note that $256$-bit codes are only used
with LSH and SIFT vectors.

\figref{fig:recall_1M_1B} shows Euclidean NN recall rates for \knn{} 
search on binary codes generated from $1$M and $1$B SIFT descriptors.  
In particular, we plot the fraction of
Euclidean $1^{st}$ nearest neighbors found, by \knn{} in $64$- and
$128$-bit LSH~\cite{Charikar02} and MLH~\cite{NorouziNIPS12} binary
codes.  As expected $128$-bit codes are more accurate, and MLH
outperforms LSH.  Note that the multi-index hashing algorithm solves
exact \knn{} search in Hamming distance; the approximation that 
reduces recall is due to the mapping from the original Euclidean space 
to the Hamming space.
To preserve the Euclidean structure in the original SIFT descriptors, 
it seems useful to use longer codes, and exploit data-dependant hash
functions such as MLH.  Interestingly, as described below, the speedup
factors of multi-index hashing on MLH codes are better than those for
LSH.

Obviously, Hamming distance computed on q-bit binary codes is an integer
between 0 and q. Thus, the nearest neighbors in Hamming distance can
be divided into subsets of elements that have equal Hamming distance
(at most q+1 subsets). Although Hamming distance does not provide a
means to distinguish between equi-distant elements, often a
re-ranking phase using Asymmetric Hamming
distance~\cite{GordoPCVPR11} or other distance measures is helpful
in practice.  Nevertheless, this paper is solely concerned with the
exact Hamming $k$NN problem up to a selection of
equi-distant elements in the top $k$ elements.

\subsection{Results}

Each experiment below involves $10^4$ queries, for which we report 
the average run-time.  
Our implementation of the linear scan baseline searches $60$ million
$64$-bit codes in just under one second on the AMD machine.  On the 
Intel machine it examines over $80$ million $64$-bit codes per second.
This is remarkably fast compared to Euclidean NN search with $128$D 
SIFT vectors. The speed of linear scan is in part due to memory caching, 
without which it would be much slower. \comment{Run-times for linear scan on 
other datasets, on both architectures, are given in 
Tables~\ref{tab:main-mih-results} and \ref{tab:main-mih-results-intel}.}

\begin{table*}
\begin{center}
\begin{small}
\begin{tabular}{|l|c|c|c|c|c|c|c|}
\cline{4-7}
\multicolumn{3}{c}{~} & \multicolumn{4}{|c|}{{\bf speedup factors for \knn{} 
{\em vs.} linear scan} } \\
\hline
dataset & \# bits & mapping & ~~~$1$-NN~~~ & ~~$10$-NN~~ & $100$-NN & $1000$-NN & linear scan\\
\hline

\hline\multirow{5}{*}{SIFT $1$B} & 
\multirow{2}{*}{$64$}  & MLH & $823$ & $757$ & $587$ & $390$ &   \multirow{2}{*}{$16.51$s }\\
&  & LSH & $781$ & $698$ & $547$ & $306$ &   \\
\cline{2-8}
& \multirow{2}{*}{$128$} & MLH & $1048$ & $675$ & $353$ & $147$ &   \multirow{2}{*}{$42.64$s }\\
&  & LSH & $747$ & $426$ & $208$ & $91$ &   \\
\cline{2-8}
& $256$ & LSH &$220$ & $111$ & $58$ & $27$ & $62.31$s \\
\hline\multirow{4}{*}{Gist $79$M} & \multirow{2}{*}{$64$}  & MLH & $401$ & $265$ & $137$ & $51$ &    \multirow{2}{*}{$1.30$s }\\
&  & LSH & $322$ & $145$ & $55$ & $18$ &    \\
\cline{2-8}
& \multirow{2}{*}{$128$}  & MLH & $124$ & $50$ & $26$ & $13$ &    \multirow{2}{*}{$3.37$s }\\
&  & LSH & $85$ & $33$ & $18$ & $9$ &   \\
\hline

\end{tabular}
\end{small}
\end{center}
\vspace*{-0.1cm}
\caption{Summary of results for nine datasets of binary codes on AMD 
Opteron Processor with 2MB L2 cache. 
The first four rows correspond to 1 billion binary codes, while the last four rows show the results for $79$ million codes. Codes are $64$, $128$, or $256$ bits long, obtained by LSH or MLH. The run-time of linear scan is reported along with the speedup factors for \knn{} with multi-index hashing.}
\label{tab:main-mih-results}
\vspace*{-0.1cm}
\end{table*}

\begin{table*}
\begin{center}
\begin{small}
\begin{tabular}{|l|c|c|c|c|c|c|c|}
\cline{4-7}
\multicolumn{3}{c}{~} & \multicolumn{4}{|c|}{{\bf speedup factors for \knn{} 
{\em vs.} linear scan} } \\
\hline
dataset & \# bits & mapping & ~~~$1$-NN~~~ & ~~$10$-NN~~ & $100$-NN & $1000$-NN & linear scan\\
\hline

\hline\multirow{5}{*}{SIFT $1$B} & 
\multirow{2}{*}{$64$}  & MLH & $573$ & $542$ & $460$ & $291$ &   \multirow{2}{*}{$12.23$s }\\
&  & LSH & $556$ & $516$ & $411$ & $237$ &   \\
\cline{2-8}
& \multirow{2}{*}{$128$} & MLH & $670$ & $431$ & $166$ & $92$ &   \multirow{2}{*}{$20.71$s }\\
&  & LSH & $466$ & $277$ & $137$ & $60$ &   \\
\cline{2-8}
& $256$ & LSH &$115$ & $67$ & $34$ & $16$ & $38.89$s \\
\hline\multirow{4}{*}{Gist $79$M} & \multirow{2}{*}{$64$}  & MLH & $286$ & $242$ & $136$ & $53$ &    \multirow{2}{*}{$0.97$s }\\
&  & LSH & $256$ & $142$ & $55$ & $18$ &    \\
\cline{2-8}
& \multirow{2}{*}{$128$}  & MLH & $77$ & $37$ & $19$ & $10$ &    \multirow{2}{*}{$1.64$s }\\
&  & LSH & $45$ & $18$ & $9$ & $5$ &   \\
\hline

\end{tabular}
\end{small}
\end{center}
\vspace*{-0.1cm}
\caption{Summary of results for nine datasets of binary codes on Intel
  Xeon Processor with 20MB L2 cache. Note that the speedup factors
  reported in this table for multi-index hashing are smaller than in
  \tableref{tab:main-mih-results}. This is due to the significant
  effect of cache size on the run-time of linear scan on the Intel
  architecture.}
\label{tab:main-mih-results-intel}
\vspace*{-0.1cm}
\end{table*}

\comment{{{
We report running time speedup factors for \knn{} using multi-index
hashing over linear scan for both AMD and Intel Architectures.  The
AMD Opteron machine is considered a more typical example of a machine
available in a research setting than the Intel machine, which has a
significantly higher cost.  One primary difference between the two
architectures is the size of each processor's L2 cache. The Intel
machine has ten times the cache of the AMD machine, which translates
into a roughly doubled speed for sequential linear scan
search. Multi-index hashing does not have a serial memory access
pattern and so cache size does not have such a pronounced
effect. Actual running times for multi-index hashing between the two
architectures agree within roughly $20$\% , with the AMD machine
sometimes being faster.  \tableref{tab:main-mih-results} and \tableref{tab:main-mih-results-intel} report speedup factors for the two
architectures. Figures \ref{64-lsh-1B} - \ref{256-lsh-1B} are
performance plots for the AMD machine.

Our multi-index hashing method solves exact $1000$-NN for a dataset of
one billion $64$-bit codes in about $50$ ms, over $300$ times faster
than linear scan (see \tableref{tab:main-mih-results}). Performance
on $1$-NN and $10$-NN are even more impressive.
}}}

\newlength\figh
\setlength\figh{5cm} 
\begin{figure*}[t]
\begin{center}
\includegraphics{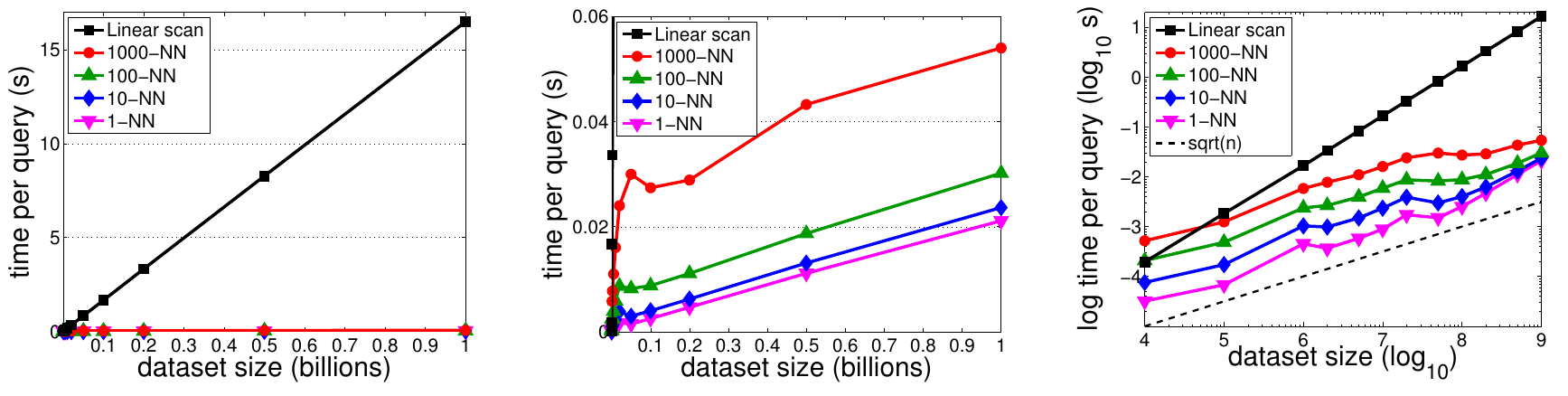}
\end{center}
\vspace*{-.25cm}
\caption{
\label{fig:64-lsh-1B}
Run-times per query for multi-index hashing with 1, 10, 100, and
1000 nearest neighbors, and a linear scan baseline on 1B 64-bit
binary codes given by LSH from SIFT. Run on an AMD Opteron processor.}

\vspace*{.4cm}

\begin{center}
\includegraphics{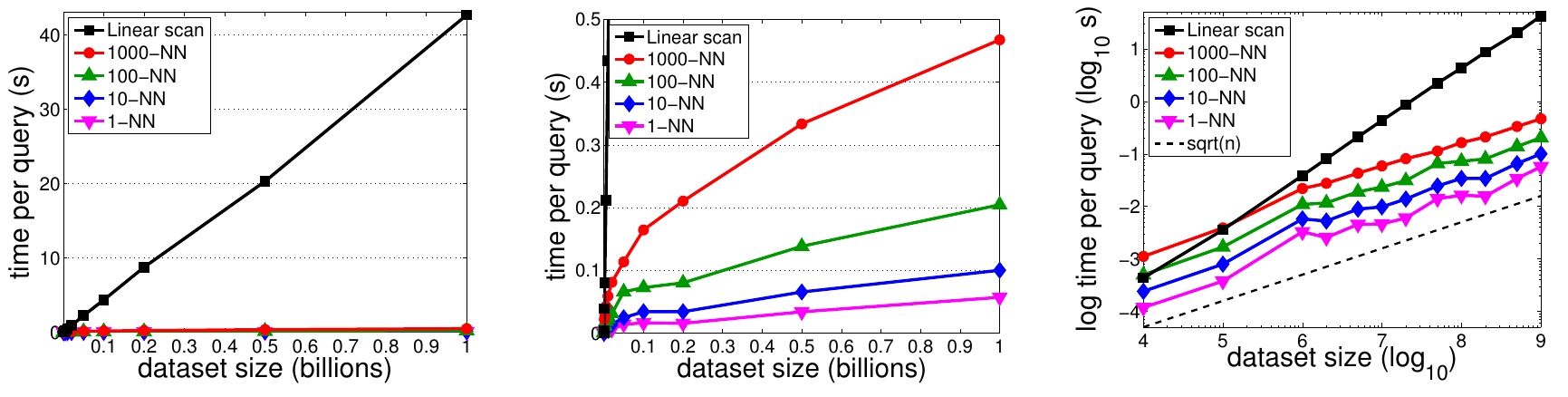}
\end{center}
\vspace*{-.25cm}
\caption{
\label{fig:128-lsh-1B}
Run-times per query for multi-index hashing with 1, 10, 100, and
1000 nearest neighbors, and a linear scan baseline on 1B 128-bit
binary codes given by LSH from SIFT. Run on an AMD Opteron processor. }

\vspace*{.4cm}

\begin{center}
\includegraphics{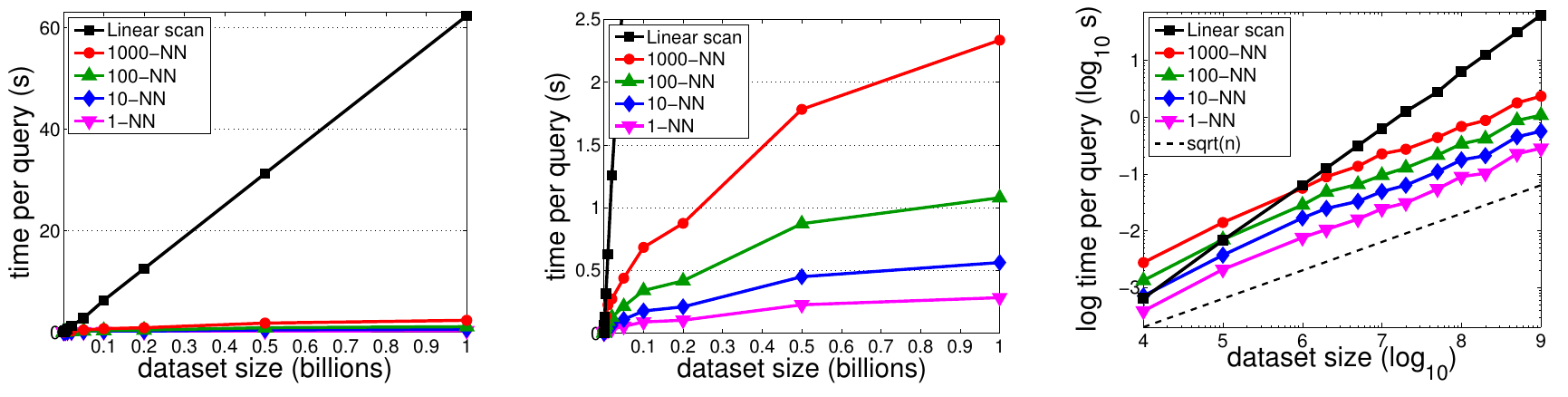}
\end{center}
\vspace*{-.2cm}
\caption{
\label{fig:256-lsh-1B}
Run-times per query for multi-index hashing with 1, 10, 100, and
1000 nearest neighbors, and a linear scan baseline on 1B 256-bit
binary codes given by LSH from SIFT. Run on an AMD Opteron processor.}
\end{figure*}

\subsection{Multi-Index Hashing {\em vs.} Linear Scan}

Tables~\ref{tab:main-mih-results} and \ref{tab:main-mih-results-intel} 
shows run-time per query for the linear scan baseline, along with speedup 
factors of multi-index hashing for different \knn{} problems and nine 
different datasets.  
Despite the remarkable speed of linear scan, the multi-index hashing
implementation is hundreds of times faster.
For example, the multi-index hashing method solves the exact $1000$-NN 
for a dataset of 1B $64$-bit codes in about $50$ ms, well over $300$ times 
faster than linear scan (see \tableref{tab:main-mih-results}). 
Performance on $1$-NN and $10$-NN are even more impressive.  
With $128$-bit MLH codes, multi-index hashing executes the 1NN search 
task over 1000 times faster than the linear scan baseline. 

The run-time of linear scan does not depend on the number of neighbors, 
nor on the underlying distribution of binary codes.  The run-time for 
multi-index hashing, however, depends on both factors.
In particular, as the desired number of NNs increases, the Hamming radius 
of the search also increases (\eg~see \figref{fig:radius-for-10-1000NN}).
This implies longer run-times for multi-index hashing.  Indeed, notice 
that going from $1$-NN to $1000$-NN on each row of the tables 
shows a decrease in the speedup factors.

The multi-index hashing run-time also depends on the distribution
of binary codes.  Indeed, one can see from
\tableref{tab:main-mih-results} that MLH code databases yield faster
run times than the LSH codes; e.g., for $100$-NN in $1$B $128$-bit
codes the speedup for MLH is $353\!\times$ {\em vs} $208\!\times$ for
LSH.  \figref{fig:radius-for-10-1000NN} depicts the histograms of
search radii needed for $1000$-NN with $1$B $128$-bit MLH and LSH
codes.  Interestingly, the mean of the search radii for MLH codes is
$19.9$ bits, while it is $19.8$ for LSH. While the means are similar
the variances are not; the standard deviations of the search radii 
for MLH and LSH are $4.0$ and $5.0$ respectively. The longer tail of 
the distribution of search radii for LSH plays an important role in 
the expected run-time. In fact, queries that require relatively 
large search radii tend to dominate the average query cost. 

\begin{figure*}
\begin{center}
\includegraphics{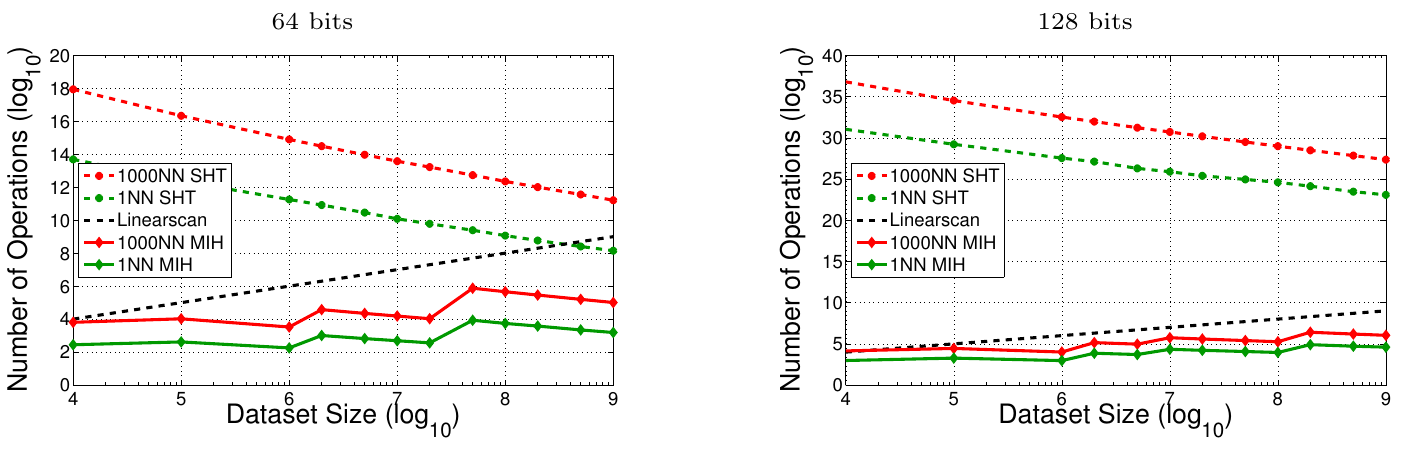}
\end{center}
\vspace*{-.2cm}
\caption{ The number of lookup operations required to solve exact
  nearest neighbor search in hamming space for LSH codes from SIFT
  features, using the simple single hash table (SHT) approach and
  multi-index hashing (MIH). Also shown is the number of Hamming
  distance comparisons required to search using linear scan.  Note
  that the axes have a logarithmic scale.  With small codes (64 bits),
  many items (1 billion) and small search distance (1 NN), it is
  conceivable that a single hash table might be faster than linear
  scan. In all other cases, a single hash table requires many orders
  of magnitude more operations than linear scan.  Note also that MIH
  will never require more operations than a single hash table - in the
  limit of very large dataset sizes, MIH will use only one hash table,
  hence MIH and SHT become equivalent.  }
\label{fig:singleHashFig}
\end{figure*}

It is also interesting to look at the multi-index hashing run-times as
a function of $\N$, the number of binary codes in the database. To
that end, \figref{fig:64-lsh-1B}, \ref{fig:128-lsh-1B}, and
\ref{fig:256-lsh-1B} depict run-times for linear scan and multi-index
\knn{} search on 64, 128, and 256-bit codes on the AMD machine. The
left two panels in each figure show different vertical scales
(since the behavior of multi-index \knn{} and linear scan are hard to
see at the same scale).  The right-most panels show the same data on
log-log axes.  First, it is clear from these plots that multi-index
hashing is much faster than linear scan for a wide range of dataset
sizes and $k$.  Just as importantly, it is evident from the log-log
plots that as we increase the database size, the speedup factors
improve. The dashed lines on the log-log plots depict $\sqrt{\N}$ (up
to a scalar constant).  The similar slope of multi-index hashing
curves with the square root curves show that multi-index hashing
exhibits sub-linear query time, even for the empirical, non-uniform
distributions of codes.

\subsection{Direct lookups with a single hash table}

An alternative to linear scan and multi-index hashing is to hash the
entire codes into a single hash table (SHT), and then use direct
hashing with each query.  As suggested in the introduction and
\figref{fig:growth-n-choose-k}, although this approach avoids the need
for any candidate checking, it may require a prohibitive number of
lookups.  Nevertheless, for sufficiently small code lengths or search
radii, it may be effective in practice.

Given the complexity associated with efficiently implementing
collision detection in large hash tables, we do not directly
experiment with the single hash table approach. Instead, we consider
the empirical number of lookups one would need, as compared to the
number of items in the database.  If the number of lookups is vastly
greater than the size of the dataset one can readily conclude that
linear scan is likely to be as fast or faster than direct indexing
into a single hash table.

Fortunately, the statistics of neighborhood sizes and required search
radii for \knn{} tasks are available from the linear scan and
multi-index hashing experiments reported above. For a given query, one
can use the $\th{k}$ nearest neighbor's Hamming distance to compute
the number of lookups from a single hash table that are required to
find all of the query's $k$ nearest neighbors.  Summed over the set of
queries, this provides an indication of the expected run-time.

\figref{fig:singleHashFig} shows the average number of lookups required
for 1-NN and 1000-NN tasks on $64$- and $128$-bit codes (from LSH on SIFT)
using a single hash table.  They are plotted as a function of the size
of the dataset, from $10^4$ to $10^9$ items.  For comparison, the
plots also show the number of database items, and the number of
lookups that were needed for multi-index hashing.  Note that
\figref{fig:singleHashFig} has logarithmic scales.

\begin{table*}
\begin{center}
\begin{small}
\begin{tabular}{|c|c|c|c|c|}
\cline{2-5} \multicolumn{1}{c}{~} & \multicolumn{4}{|c|}{{\bf
    optimized speedup vs. linear scan (consecutive, \% improvement)}
} \\ \hline \# bits & ~~~$1$-NN~~~ & ~~$10$-NN~~ & $100$-NN &
$1000$-NN \\ \hline

\hline $64$ & $788$ ($781$, $1$\%) & $750$ ($698$, $7$\%) & $570$ ($547$, $4$\%) & $317$ ($306$, $4$\%) \\
\hline $128$ & $826$ ($747$, $10$\%) & $472$ ($426$, $11$\%) & $237$ ($208$, $14$\%) & $103$ ($91$, $12$\%) \\
\hline $256$ & $284$ ($220$, $29$\%) & $138$ ($111$, $25$\%) & $68$ ($58$, $18$\%) & $31$ ($27$, $18$\%) \\

\hline
\end{tabular}
\end{small}
\end{center}
\caption{Empirical run-time improvements from optimizing substrings
  {\em vs.} consecutive substrings, for 1 billion LSH codes from SIFT
  features (AMD machine).  speedup factors vs. linear scan are shown
  with optimized and consecutive substrings, and the percent
  improvement. All experiments used 10M codes to compute the
  correlation between bits for substring optimization and all results
  are averaged over 10000 queries each.}
\label{tab:optimal-substring-results-multi}
\vspace*{-0.3cm}
\end{table*}

It is evident that with a single hash table the number of lookups is
almost always several orders of magnitude larger than the number of
items in the dataset.  And not surprisingly, this is also several
orders of magnitude more lookups than required for multi-index
hashing.  Although the relative speed of a lookup operation compared
to a Hamming distance comparison, as used in linear scan, depends on the
implementation, there are a few important considerations. Linear scan
has an exactly serial memory access pattern and so can make very
efficient use of cache, whereas lookups in a hash table are inherently
random.  Furthermore, in any plausible implementation of a single hash
table for 64 bit or longer codes, there will be some penalty for
collision detection.

As illustrated in \figref{fig:singleHashFig}, the only cases where 
a single hash table might potentially be more efficient than linear scan 
are with very small codes (64 bits or less), with a large dataset 
(1 billion items or more), and a small search distances (e.g., for 1-NN).  
In all other cases, linear scan requires orders of magnitude fewer 
operations. With any code length longer than 64 bits, a single hash 
table approach is completely infeasible to run, requiring upwards of 
15 orders of magnitude more operations than linear scan for 128-bit codes.


\subsection{Substring Optimization}

The substring hash tables used above have been formed by simply
dividing the full codes into disjoint and consecutive sequences
of bits.  For LSH and MLH, this is equivalent to randomly
assigning bits to substrings.

It natural to ask whether further gains in efficiency are possible by
optimizing the assignment of bits to substrings.  In particular, by
careful substring optimization one may be able to maximize the
discriminability of the different substrings.  In other words, while
the radius of substring searches and hence the number of lookups is
determined by the desired search radius on the full codes, and will
remain fixed, by optimizing the assignment of bits to substrings one
might be able to reduce the number of candidates one needs to
validate.

To explore this idea we considered a simple method in which bits are
assigned to substrings one at a time in a greedy fashion based on the
correlation between the bits. We initialize the substrings greedily. A
random bit is assigned to the first substring. Then, a bit is assigned
to substring $j$, which is maximally correlated with the bit assigned
to substring $j-1$. Next, we iterate over the substrings, and assign
more bits to them, one at a time. An unused bit is assigned to
substring $j$, if the maximum correlation between that bit and other
bits already assigned to substring $j$ is minimal. This approach
significantly decreases the correlation between bits within a single
substring. This should make the distribution of codes within
substrings buckets more uniform, and thereby lower the number of
candidates within a given search radius.  Arguably, a better approach
consists of maximizing the entropy of the entries within each
substring hash table, thereby making the distribution of substrings as
uniform as possible. However, this entropic approach is left to future
work.

\begin{figure*}[t]
\begin{center}
\includegraphics{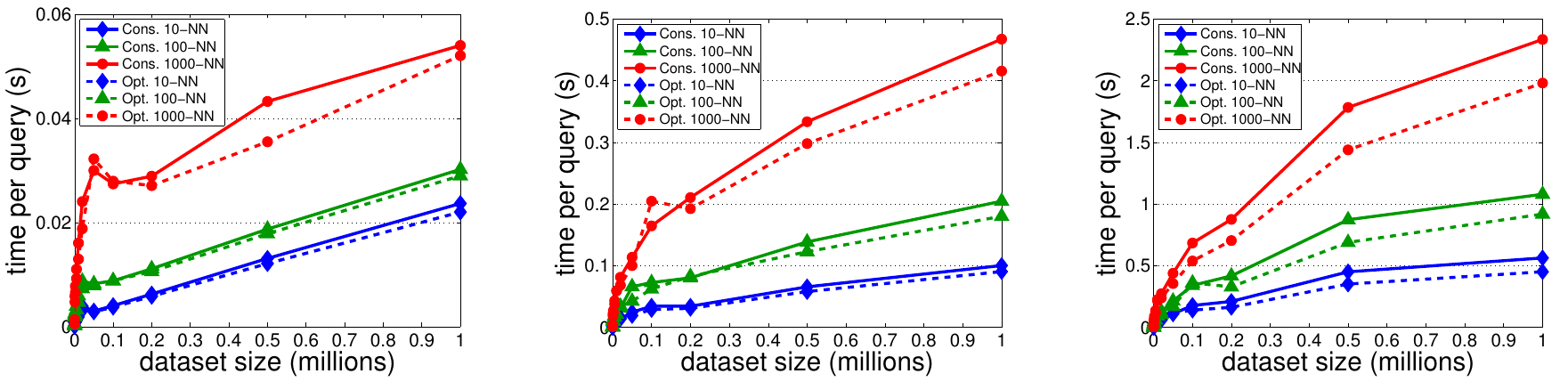}
\end{center}
\vspace*{-.3cm}
\caption{
\label{fig:optimal-substring-plots}
Run-times for multi-index-hashing using codes from LSH on SIFT
features with consecutive (solid) and optimized (dashed)
substrings. From left to right: 64-bit, 128-bit, 256-bit codes, run on
the AMD machine.}
\end{figure*}

The results obtained with the correlation-based greedy algorithm show
that optimizing substrings can provide overall run-time reductions on
the order of $20\%$ against consecutive substrings for some
cases. \tableref{tab:optimal-substring-results-multi} displays the
improvements achieved by optimizing substrings for different codes
lengths and different values of $k$. Clearly, as the code length
increases, substring optimization has a bigger impact.
\figref{fig:optimal-substring-plots} shows the run-time behavior of
optimized substrings as a function of dataset size.

\begin{table*}
\begin{center}
\begin{small}
\begin{tabular}{|@{\hspace*{.1cm}}r@{\hspace*{.2cm}}|@{\hspace*{.2cm}}c@{\hspace*{.2cm}}|c|c|c|@{\hspace*{.1cm}}c@{\hspace*{.1cm}}|@{\hspace*{.1cm}}c@{\hspace*{.1cm}}|c|@{\hspace*{.1cm}}c@{\hspace*{.1cm}}|@{\hspace*{.1cm}}c@{\hspace*{.1cm}}|c|@{\hspace*{.1cm}}c@{\hspace*{.1cm}}|@{\hspace*{.1cm}}c@{\hspace*{.1cm}}|c|}
\hline
& $\N$             & $~10^4~$ & $~10^5~$ & $~10^6~$ & $2 \times 10^6$ & $5 \times 10^6$ & $~10^7~$ & $2 \times 10^7$ & $5 \times 10^7$ & $~10^8~$ & $2 \times 10^8$ & $5 \times 10^8$ & $~10^9~$ \\
\hline
\multirow{2}{*}{ $q = 64$} & $\M$             & $5$    & $4$    & $4$    & $3$             & $3$             & $3$    & $3$            & $2$             & $2$    & $2$             & $2$             & $2$    \\
                          & $\B / \log_2 \N$ & $4.82$ & $3.85$ & $3.21$ & $3.06$          & $2.88$          & $2.75$ & $2.64$         & $2.50$          & $2.41$ & $2.32$          & $2.21$          & $2.14$ \\
\hline
\multirow{2}{*}{ $q = 128$} & $\M$             & $10$ & $8$ & $8$ & $6$ & $6$ & $5$ & $5$ & $5$ & $5$ & $4$ & $4$ & $4$ \\
                          & $\B / \log_2 \N$ & $9.63$ & $7.71$ & $6.42$ & $6.12$ & $5.75$ & $5.50$ & $5.28$ & $5.00$ & $4.82$ & $4.64$ & $4.43$ & $4.28$ \\
\hline
\multirow{2}{*}{ $q = 256$} & $\M$             & $19$ & $15$ & $13$ & $12$ & $11$ & $11$ & $10$ & $10$ & $10$ & $9$ & $9$ & $8$ \\
                          & $\B / \log_2 \N$ & $19.27$ & $15.41$ & $12.84$ & $12.23$ & $11.50$ & $11.01$ & $10.56$ & $10.01$ & $9.63$ & $9.28$ & $8.86$ & $8.56$ \\
\hline
\end{tabular}

\end{small}
\end{center}
\vspace*{-0.1cm}
\caption{Selected number of substrings used for the 
experiments, as determined by cross-validation, \vs~the suggested 
number of substrings based on the heuristic $\B\,/\log_2 \N$.  }
\label{tab:selected-m}
\vspace*{-0.1cm}
\end{table*}

\section{Implementation details}
\label{sec:details}

Our implementation of multi-index hashing is publicly available 
at \cite{mih}.  Nevertheless, for the interested reader we 
describe some of the important details here.

As explained above, the algorithm hinges on hash tables built on
disjoint $\s$-bit substrings of the binary codes. We use direct 
address tables for the substring hash tables because the substrings 
are usually short ($s \le 32$). Direct address tables explicitly
allocate memory for $2^s$ buckets and store all data points associated 
with each substring in its corresponding bucket. There is a one-to-one 
mapping between buckets and substrings, so no time is spent on 
collision detection.

One could implement direct address tables with an array of $2^s$
pointers, some of which may be null (for empty buckets).  On a
$64$-bit machine, pointers are $8$ bytes long, so just storing an
empty address table for $s = 32$ requires $32\,$GB (as done
in~\cite{NorouziPF12}).  For greater efficiency here, we use sparse
direct address tables by grouping buckets into subsets of $32$
elements. For each bucket group, a $32$-bit binary vector encodes
whether each bucket in the group is empty or not.  Then, a single
pointer per group is used to point to a single resizable
array that stores the data points associated with that bucket group.
Data points within each array are ordered by their bucket index.  To
facilitate fast access, for each non-empty bucket we store the index
of the beginning and the end of the corresponding segment of the
array.  Compared to the direct address tables in~\cite{NorouziPF12},
for $s = 32$, and bucket groups of size $32$, an empty address table
requires only $1.5\,$GB.  Also note that accessing elements in any
bucket of the sparse address table has a worst case run-time of
$O(1)$.

\vspace*{0.1cm}
\noindent 
{\em Memory Requirements:} We store one $64$-bit pointer for each
bucket group, and a $32$-bit binary vector to encode whether buckets 
in a group are empty; this entails $2^{(s-5)}\cdot(8+4)$ bytes for an 
empty $s$-bit hash table ($s \ge 5$), or $1.5\,$GB when $s = 32$.  
Bookkeeping for each resizable array entails $3$ $32$-bit integers. 
In our experiments, most bucket groups have at least one non-empty 
bucket.  Taking this into account, the total storage for an $s$-bit 
address table becomes $2^{(s-5)}\cdot24$ bytes ($3\,$GB for $s = 32$).

For each non-empty bucket within a bucket group, we store a $32$-bit
integer to indicate the index of the beginning of the segment of the
resizable array corresponding to that bucket. The number of non-empty
buckets is at most $\M\min(\N,2^\s)$, where $\M$ is the number of hash
tables, and $\N$ is the number of codes. Thus we need an extra 
$\M\min(\N,2^\s)\cdot4$ bytes. For each data point per hash table we 
store an ID to reference the full binary code; each ID is $4$ bytes 
since $\N \le 2^{32}$ for our datasets.  This entails $4\M\N$ bytes. 
Finally, storing the full binary codes themselves requires $\N\M\s/8$ 
bytes, since $\B = \M\s$.

The total memory cost is $\M2^{(\s-5)}24 + \M \min(\N,
2^\s)4 + 4\M\N + \N\M\s/8$ bytes.  For $s= \log_2 n$, this cost 
is $O(nq)$.  For 1B $64$-bit codes, and $m=2$ hash tables
($32$ bits each), the cost is $28\,$GB.  For $128$-bit and
$256$-bit codes our implementation requires $57\,$GB and $113\,$GB.
Note that the last two terms in the memory cost for storing IDs and
codes are irreducible, but the first terms can be reduced in a more
memory efficient implementation.

\vspace*{0.1cm}
\noindent
{\em Duplicate Candidates: } When retrieving candidates from the $\M$
substring hash tables, some codes will be found multiple times.  To
detect {\em duplicates}, and discard them, we allocate one bit-string
with $\N$ bits.  When a candidate is found we check the corresponding
bit and discard the candidate if it is marked as a duplicate.  Before
each query we initialize the bit-string to zero. In practice this has
negligible run-time.  In theory clearing an $\N$-bit vector requires
$O(\N)$, but there exist ways to initialize an $O(\N)$ array
  in constant time.

\vspace*{0.1cm}
\noindent
{\em Hamming Distance: } To compare a query with a candidate
(for multi-index search or linear scan), we compute Hamming distance
on the full $\B$-bit codes, with one \verb|xor| operation for every 64
bits, followed by a pop count to tally the ones.  We used the built-in
GCC function \verb|__builtin_popcount| for this purpose.

\vspace*{0.1cm}
\noindent
{\em Number of Substrings: } The number of substring hash tables we
use is determined with a hold-out validation set of database entries.
From that set we estimate the running time of the algorithm for
different choices of $\M$ near $\B\,/\log_2 \N$, and select the
$\M$ that yields the minimum run-time.  As shown in
\tableref{tab:selected-m} this empirical value for $\M$ is usually the
closest integer to $\B\,/\log_2 \N$.

\comment{{
\david{do we need the rest of this?}
There are more details involved in the
implementation that we will disclose by publishing the code online
after publication.

Given a desired number of neighbors, $K$, for each
validation entry we compute the required search radius per substring,
and analytically compute the number of lookups for each $\M$. etc.
}}

\vspace*{0.1cm}
\noindent
{\em Translation Lookaside Buffer and Huge Pages:} Modern processors
have an on-chip cache that holds a lookup table of memory addresses,
for mapping virtual addresses to physical addresses for each running 
process.  Typically, memory is split into 4KB pages, and a process 
that allocates memory is given pages by the operating system. The 
Translation Lookaside Buffer (TLB) keeps track of these pages.  For 
processes that have large memory footprints (tens of GB), the number 
of pages quickly overtakes the size of the TLB (typically about 1500 
entries).  For processes using random memory access this means that 
almost every memory access produces a {\em TLB miss} - the requested 
address is in a page not cached in the TLB, hence the TLB entry must 
be fetched from slow RAM before the requested page can be accessed.  
This slows down memory access, and causes volatility in run-times 
for memory-access intensive processes.

To avoid this problem, we use the \verb|libhugetlbfs| Linux library.
This allows the operating system to allocate {\em Huge Pages} (2MB
each) rather than 4KB pages. This reduces the number of pages and the
frequency of TLB misses, which improves memory access speed, and
reduces run-time volatility. The increase in speed of multi-index
hashing compared to the results reported in \cite{NorouziPF12} is
partly attributed to the use of libhugetlbfs.





\comment{{{
\section{Discussion}
\david{If these subsections are short enough we could append them to 
the end of the conclusions in the context of future work.}

\subsection{Approximate Algorithms}
While outside the scope of this paper, it would also be interesting
to consider approximate methods based on multi-index hashing.
There are several ways that one can generate approximate results.

And it would be natural to compare such methods against other state-of-the-art
approximate methods, such as binary LSH.  Such comparisons are somewhat
more difficult than for exact methods since one must taken into account
not only the storage and run-time costs, but also some measure of the 
cost of errors (usually in terms of recall and precision).

\subsection{Asymmetric Hamming Distance}
\david{Discuss AH distance and why it's useful, and include the idea 
that re-ranking is a simple but effective approach. There may be other methods
that are much more efficient, but we also leave this to future work.}
}}}

\section{Conclusion}

This paper describes a new algorithm for exact nearest neighbor search
on large-scale datasets of binary codes.  The algorithm is a form of
multi-index hashing that has provably sub-linear run-time behavior for
uniformly distributed codes.  It is storage efficient and easy to
implement.  We show empirical performance on datasets of binary codes
obtained from $1$ billion SIFT, and $80$ million Gist features.  With
these datasets we find that, for $64$-bit and $128$-bit codes, our
new multi-index hashing implementation is often more than two orders of
magnitude faster than a linear scan baseline.

While the basic algorithm is developed in this paper there are 
several interesting avenues for future research. For example
our preliminary research shows that $\log_2 \N$ is a good choice 
for the substring length, and it should be possible to formulate a 
sound mathematical basis for this choice.  The assignment of 
bits to substrings was shown to be important above, however the
algorithm used for this assignment is clearly suboptimal.  It is 
also likely that different substring lengths might be useful for 
the different hash tables.  

Our theoretical analysis proves sub-linear run-time behavior of the
multi-index hashing algorithm on uniformly distributed codes, when
search radius is small. Our experiments demonstrate sub-linear
run-time behavior of the algorithm on real datasets, while the
binary code in our experiments are clearly not uniformly
distributed\footnote{In some of our experiments with $1$ Billion
binary codes, tens of thousands of codes fall into the same bucket of
$32$-bit substring hash tables. This is extremely unlikely with
uniformly distributed codes.}. Bridging the gap between 
theoretical analysis and empirical findings for the proposed
algorithm remains an open problem. In particular, we are interested 
in more realistic assumptions on the binary codes, which still 
allow for theoretical analysis of the algorithm.

While the current paper concerns exact nearest-neighbor tasks, it would 
also be interesting to consider approximate methods based on the same 
multi-index hashing framework.  Indeed there are several ways that one 
could find approximate rather than the exact nearest neighbors for a given 
query. For example, one could stop at a given radius of search, even though
$k$ items may not have been found.  Alternatively, one might search until 
a fixed number of unique candidates have been found, even though all 
substring hash tables have not been inspected to the necessary radius,  
Such approximate algorithms have the potential for even greater efficiency, 
and would be the most natural methods to compare to most existing methods 
which are approximate, such as binary LSH.  That said, such comparisons are 
more difficult than for exact methods since one must taken into account not 
only the storage and run-time costs, but also some measure of the cost of 
errors (usually in terms of recall and precision).

Finally, recent results have shown that for many datasets in which the
binary codes are the result of some form of vector quantization, an
asymmetric Hamming distance is
attractive \cite{GordoPCVPR11,jegouPAMI11}.  In such methods, rather
than converting the query into a binary code, one directly compares a
real-valued query to the database of binary codes.  The advantage is
that the quantization noise entailed in converting the query to a
binary string is avoided and one can more accurately using distances
in the binary code space to approximate the desired distances in the
feature space of the query.  One simple way to do this is to use
multi-index hashing and then only use an asymmetric distance when
culling candidates.  The potential for more interesting and effective
methods is yet another promising avenue for future work.

\ifCLASSOPTIONcompsoc
  \section*{Acknowledgments}
\else
  \section*{Acknowledgment}
\fi

This research was financially supported in part by NSERC Canada, the
GRAND Network Centre of Excellence, and the Canadian Institute for
Advanced Research (CIFAR).  The authors would also like to thank 
Mohamed Aly, Rob Fergus, Ryan Johnson, Abbas Mehrabian, and Pietro Perona 
for useful discussions about this work.

\ifCLASSOPTIONcaptionsoff
  \newpage
\fi



%

\bibliographystyle{ieee}
\bibliography{myrefs}

\end{document}